%% file: sample-authordraft.tex
\documentclass[conference, 10pt]{IEEEtran}
\IEEEoverridecommandlockouts

\usepackage{url}
\usepackage{cite}
\usepackage{amsmath,amssymb,amsfonts}
\usepackage{algorithmic}
\usepackage{graphicx}
\usepackage{textcomp}
\usepackage{xcolor}
\def\BibTeX{{\rm B\kern-.05em{\sc i\kern-.025em b}\kern-.08em
    T\kern-.1667em\lower.7ex\hbox{E}\kern-.125emX}}

\usepackage{enumitem}

\usepackage{romannum}
\usepackage{textcomp}
\usepackage{tabularx}
\usepackage{diagbox}
\usepackage{multirow}
\usepackage{hhline}

\usepackage{bbding}
\usepackage{pifont}
\usepackage{wasysym}

\usepackage{cleveref}
\crefformat{section}{\S#2#1#3}
\crefformat{subsection}{\S#2#1#3}
\crefformat{subsubsection}{\S#2#1#3}

\usepackage{listings}
\usepackage{color}
\definecolor{dkgreen}{rgb}{0,0.6,0}
\definecolor{gray}{rgb}{0.5,0.5,0.5}
\definecolor{mauve}{rgb}{0.58,0,0.82}
\definecolor{backcolour}{rgb}{0.95,0.95,0.92}
\usepackage{graphics}
\usepackage{caption}
\usepackage{subcaption}
\graphicspath{{./figures/}}

\usepackage{amsmath}
\usepackage{mathrsfs}

\begin{document}

\title{Poster: Making Edge-assisted LiDAR Perceptions Robust to Lossy Point Cloud Compression}

\author
{
    \IEEEauthorblockN{
      Jin Heo\IEEEauthorrefmark{1}\IEEEauthorrefmark{3},
      Gregoire Phillips\IEEEauthorrefmark{2}\IEEEauthorrefmark{4},
      Per-Erik Brodin\IEEEauthorrefmark{2}\IEEEauthorrefmark{5},
      Ada Gavrilovska\IEEEauthorrefmark{1}\IEEEauthorrefmark{6}}
    \IEEEauthorblockA{\IEEEauthorrefmark{1} Georgia Institute of Technology, Atlanta, Georgia, USA}
    \IEEEauthorblockA{\IEEEauthorrefmark{2} Ericsson Research, Santa Clara, California, USA}
    Email: \IEEEauthorrefmark{3}jheo33@gatech.edu,
           \IEEEauthorrefmark{4}greg.phillips@ericsson.com,
           \IEEEauthorrefmark{5}per-erik.brodin@ericsson.com,
           \IEEEauthorrefmark{6}ada@cc.gatech.edu
}





\pagenumbering{arabic}

\newcommand{\jin}[1]{\textcolor{blue}{JH: #1}}
\newcommand{\ada}[1]{\textcolor{red}{AG: #1}}
\newcommand{\blue}[1]{\textcolor{blue}{#1}}
\newcommand{\red}[1]{\textcolor{red}{#1}}
\newcommand{\green}[1]{\textcolor{green}{#1}}
\newcommand{\eg}{\emph{e.g.}}
\newcommand{\mul}{$\times$}

\newenvironment{tightitemize}%
 {\begin{list}{$\bullet$}{%
 		\setlength{\leftmargin}{10pt}
        \setlength{\itemsep}{0pt}%
        \setlength{\parsep}{0pt}%
        \setlength{\topsep}{0pt}%
        \setlength{\parskip}{0pt}%
        }%
 }%
{\end{list}}

\newcommand{\specialcell}[2][c]{%
  \begin{tabular}[#1]{@{}c@{}}#2\end{tabular}}

\maketitle
\thispagestyle{plain}
\pagestyle{plain}


\input{sections/abstract.tex}

\input{sections/introduction.tex}
\input{sections/ri_interpolation.tex}

\input{sections/summary.tex}

\bibliographystyle{IEEEtran}
\bibliography{sample-base}

\end{document}

%% file: sections/abstract.tex
\begin{abstract}
Real-time light detection and ranging (LiDAR) perceptions, \eg, 3D object detection and simultaneous localization and mapping are computationally intensive to mobile devices of limited resources and often offloaded on the edge.
Offloading LiDAR perceptions requires compressing the raw sensor data, and lossy compression is used for efficiently reducing the data volume.
Lossy compression degrades the quality of LiDAR point clouds, and the perception performance is decreased consequently.
In this work, we present an interpolation algorithm improving the quality of a LiDAR point cloud to mitigate the perception performance loss due to lossy compression.
The algorithm targets the range image (RI) representation of a point cloud and interpolates points at the RI based on depth gradients.
Compared to existing image interpolation algorithms, our algorithm shows a better qualitative result when the point cloud is reconstructed from the interpolated RI.
With the preliminary results, we also describe the next steps of the current work.
\end{abstract}

%% file: sections/introduction.tex
\section{Introduction}

A light detection and ranging (LiDAR) sensor enables 3D sensing capability.
It emits laser beams and generates a point cloud by measuring the time taken to receive the reflected light pulses.
Since a LiDAR sensor allows using 3D environmental information and is more robust to light and weather conditions than 2D camera sensors~\cite{khader2020introduction}, it has been widely applied to self-driving cars and robotics.
In the past, LiDAR sensors were so costly and big in size that few device platforms leveraged them~\cite{expensive}.
Recently, LiDAR sensors are more available to mobile devices because they are getting smaller, cost-effective, and low-power~\cite{velabit, realsense, applelidar}.
So, there are more opportunities for mobile devices to utilize this 3D sensing capability in diverse use cases, \eg, extended reality (XR) and 3D reconstruction.

While a LiDAR sensor is getting prevalent on mobile devices, its usage for real-time perceptions such as 3D object detection and simultaneous localization and mapping is restricted due to the prohibitive computational costs of LiDAR perception algorithms~\cite{simon2019complexer, simony2018complex, ye2020hvnet, yang2018pixor, zhang2020polarnet}.
In this situation, edge computing can be a technology for enabling computationally intensive LiDAR perceptions to mobile users.
Commodity devices can offload LiDAR perceptions on the edge, and offloading LiDAR perceptions ensures lower processing time and cost for the perception algorithms~\cite{chen2018marvel, chen2015glimpse, liu2019edge}.

When running the LiDAR perceptions on the edge in real-time, an efficient point cloud compression (PCC) method is necessary because of the large volume of raw point clouds.
Additionally, the PCC method should be lightweight to operate on mobile devices and low latency for the latency-performance tradeoff of real-time perceptions; the higher end-to-end latency causes larger discrepancies between the perception result and the real-world environment~\cite{li2020towards}.
In our recent work, we showed that existing PCC methods~\cite{draco, feng2020real, rusu20113d, tu2019real} are hardly suitable for remote real-time perceptions and presented a fast and lightweight PCC method, FLiCR~\cite{flicr2022}.

Although FLiCR achieved the low-latency 
and efficiency requirements, it compromised the data quality due to use of lossy compression, and caused the perception performance loss.
To mitigate this perception performance loss, 
we are developing a lightweight and low-latency interpolation algorithm to restore the lost points in a point cloud.
Our algorithm is based on the range image (RI) representation of a point cloud, which maps 3D points into a 2D frame.
Since manipulations at the 2D frame can cause unexpected distortions in 3D space, the  interpolation algorithm needs to be specialized for RIs of LiDAR point clouds.

%% file: sections/ri_interpolation.tex
\section{Range Image Interpolation and Preliminary Results}

A LiDAR point cloud is an unstructured point cloud having an arbitrary number of 3D points.
The existing PCC methods convert the raw point cloud into the
intermediate representations (IR), \eg, range image (depth map),
octree, k-d tree, and mesh, and compress the IRs.
Lossy compression methods achieve a higher compression ratio than lossless compression by reducing the level of detail (LoD) at the IR level and losing points in the point cloud.
Among the IRs, the range image (RI) has a low-latency benefit with the simplicity of its conversion process~\cite{flicr2022} as it is generated by converting the points in the 3D Cartesian coordinates to the spherical coordinates and mapping the converted points into a 2D image.
Each pixel of a RI has a depth value.
By presenting a point cloud as an image, it becomes possible to leverage the existing image-processing techniques.
Our work is motivated by the observation that the pixel interpolation at the RI level generates the interpolated points in 3D space, and it can be used to relieve the performance degradation of LiDAR perceptions on the edge by restoring the lost points by lossy compression.

\begin{figure}[h]
  \centering
  \begin{subfigure}[t]{0.35\textwidth}
    \includegraphics[width=\textwidth]{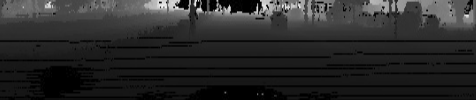}
    \caption{\small A reference RI of a point cloud.}
    \label{fig:orig}
  \end{subfigure}\hspace{0.02\textwidth}
  \begin{subfigure}[t]{0.35\textwidth}
    \includegraphics[width=\textwidth]{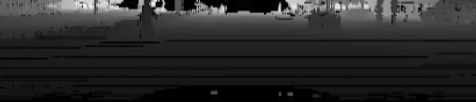}
    \caption{\small An upscaled RI by our interpolation.}
    \label{fig:ours}
  \end{subfigure}
  \begin{subfigure}[t]{0.35\textwidth}
    \includegraphics[width=\textwidth]{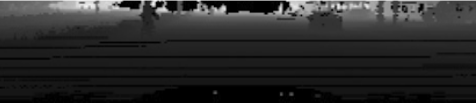}
    \caption{\small An upscaled RI by bilinear interpolation.}
    \label{fig:bilinear}
  \end{subfigure}
  \caption {\small The reference and interpolated RIs of a point cloud in the KITTI dataset.}
  \label{fig:ris}
\end{figure}

By using the existing image interpolation algorithms, we upscale a RI and reconstruct point clouds from the upscaled RIs.
Figure~\ref{fig:ris} shows the reference and interpolated RIs.
The reference RI of 1024\mul64 is from a point cloud in the KITTI dataset~\cite{geiger2013vision}.
Figure~\ref{fig:bilinear} is a part of the upscaled RI (2048\mul64) by bilinear interpolation.
Along with bilinear interpolation, we upscale the reference RI with bicubic and Lanczos interpolations.
For the interpolated RIs, we measure the image quality metrics of DSS~\cite{balanov2015image}, FSIM~\cite{zhang2011fsim}, SSIM~\cite{ssim}, VSI~\cite{zhang2014vsi}, and SR-SIM~\cite{zhang2012sr}.
Table~\ref{tab:imgmetrics} shows the image quality results of the interpolated RIs with respect to the reference RI.
The results show the interpolated RIs are objectively good-quality images to the reference RIs.

\begin{figure}[h]
  \centering
  \begin{subfigure}[t]{0.38\textwidth}
    \includegraphics[width=\textwidth]{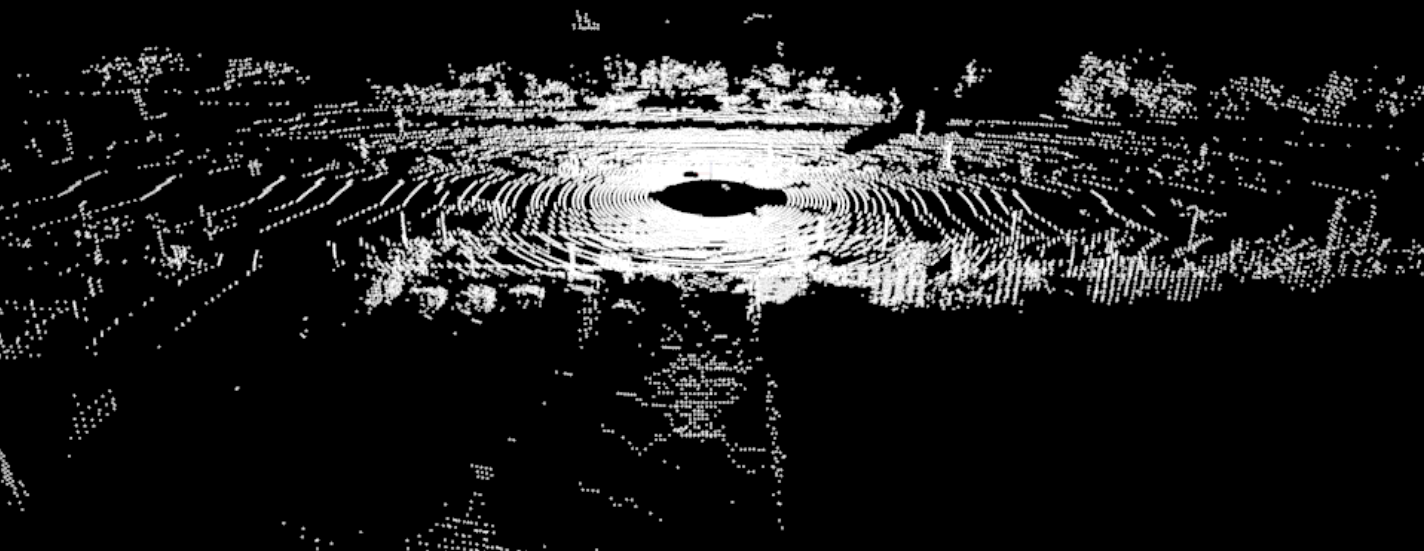}
    \caption{\small The reconstructed point cloud from an original RI.}
    \label{fig:orig_recon}
  \end{subfigure}\hspace{0.02\textwidth}
  \begin{subfigure}[t]{0.38\textwidth}
    \includegraphics[width=\textwidth]{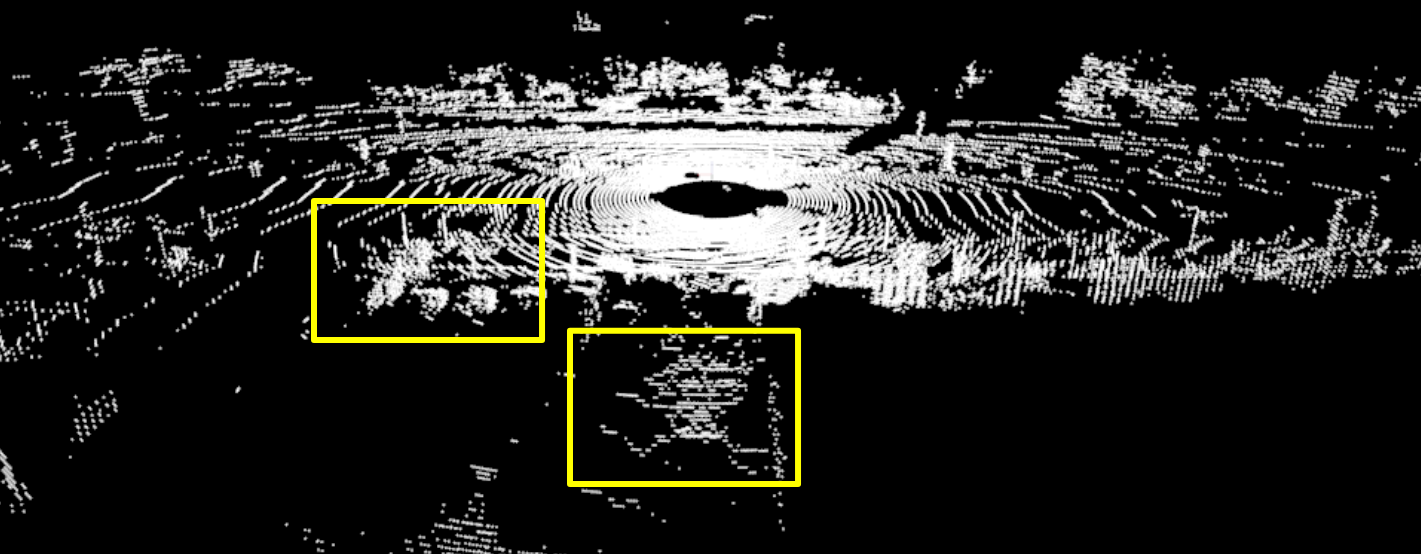}
    \caption{\small The reconstructed point cloud from an upscaled RI by our interpolation.}
    \label{fig:our_recon}
  \end{subfigure}\hspace{0.02\textwidth}
  \begin{subfigure}[t]{0.38\textwidth}
    \includegraphics[width=\textwidth]{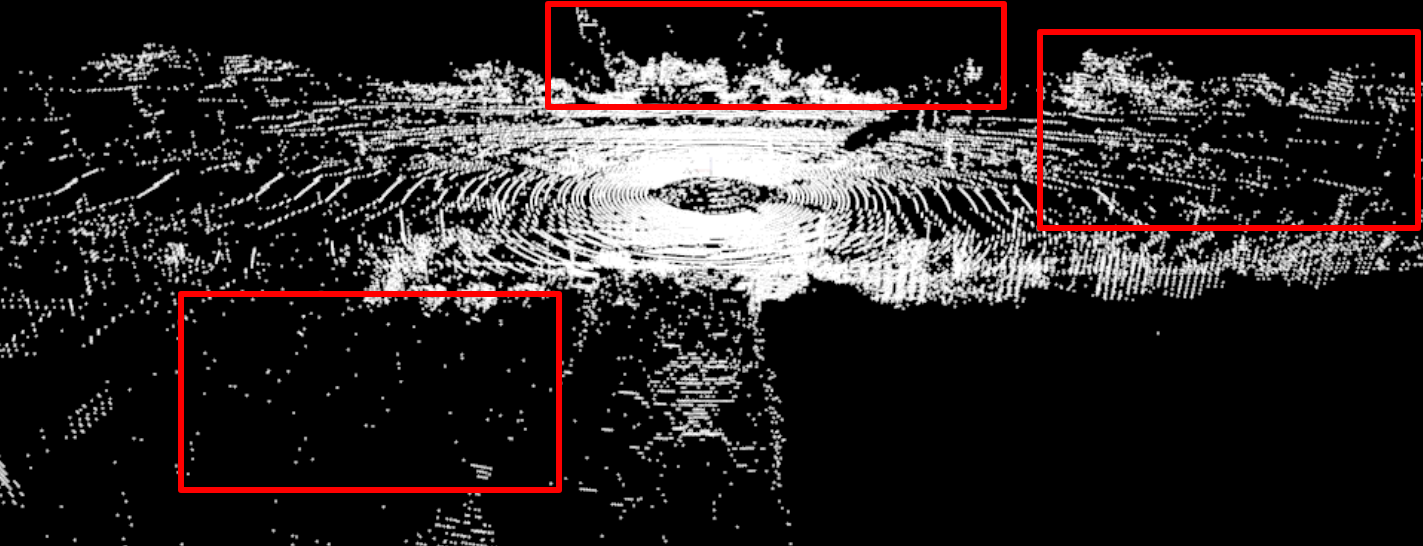}
    \caption{\small The reconstructed point cloud from an upscaled RI by bilinear interpolation.}
    \label{fig:bi_recon}
  \end{subfigure}
  \caption {\small The reconstructed point clouds from upscaled RIs.}
  \label{fig:recon}
\end{figure}

Although those RIs of high image quality scores are expected to generate high-quality point clouds, the effectiveness of the 2D image interpolations is not translated into high-quality 3D point clouds.
Figure~\ref{fig:recon} shows the point clouds reconstructed from the original and upscaled RIs.
Compared to the point cloud from the original RI (Figure~\ref{fig:orig_recon}), the point cloud from the upscaled RI by bilinear interpolation (Figure~\ref{fig:bi_recon}) has many noisy points (red boxes).
The interpolation algorithms including bilinear, bicubic, and Lanczos interpolations, leverage the near-pixel information of an interpolating pixel to put a proper value, and these noisy points are caused by the interpolating operation that blindly utilizes the near pixels; the pixels neighboring in a 2D frame can be placed far away in the 3D space or come from empty spaces.

The interpolation algorithm for RIs should be designed with awareness of the RI characteristics and how its operation in the 2D space affects the 3D point cloud.
We are developing an interpolation algorithm specialized for RIs.
Our algorithm consists of two phases: window exploration and interpolation.
Our algorithm windows a RI, and the window exploration iterates all windows of a RI and finds interpolating places within a window.
In the window exploration phase, our algorithm calculates the depth gradients and interpolating values.
When the gradients are calculated, our algorithm identifies empty pixels and invalidates the gradients between object and empty pixels.
Then, in the window interpolation phase, our algorithm interpolates pixels within a window based on the information from the exploration phase.
The interpolation policy can be set; among the possible places in a window, the policy prioritizes the places in ascending or descending order of the depth values for the interpolation priority among near or far-away objects.

Figure~\ref{fig:ours} shows the RI with our interpolation algorithm.
As shown in Table~\ref{tab:imgmetrics}, the interpolated RI with our algorithm shows lower image quality scores than the existing interpolation algorithms.
However, when the point cloud is reconstructed from the RI with our algorithm in Figure~\ref{fig:our_recon}, the interpolated pixels are effectively translated into 3D points and densify the 3D object shapes (yellow boxes), not as noisy points.

\begin{table}[htbp]
  \caption{The image quality results of the RIs interpolated by different algorithms to the reference RI (Max 1.0).}
  \begin{center}
    \scalebox{0.8}{
    \begin{tabular}{ |c|c|c|c|c|c| }
    \hline
             & DSS~\cite{balanov2015image}   & FSIM~\cite{zhang2011fsim}         & SSIM~\cite{ssim}   & VSI~\cite{zhang2014vsi}      & SR-SIM~\cite{zhang2012sr} \\ \hline
      \textbf{Ours} & \textbf{0.92} &  \textbf{0.93}  & \textbf{0.93}   & \textbf{0.97} & \textbf{0.94}  \\ \hline
     Bilinear      & 0.99 &  0.98  & 0.98   & 0.99 & 0.98  \\ \hline
     Lanczos       & 0.95 &  0.95  & 0.96   & 0.98 & 0.96  \\ \hline
     Bicubic       & 0.95 &  0.95  & 0.96   & 0.98 & 0.97  \\ \hline
  \end{tabular}
  }
  \label{tab:imgmetrics}
  \end{center}
\end{table}

%% file: sections/summary.tex
\section{Summary and Next Steps}

We motivate the need for RI interpolation techniques to compensate for
the loss of perception due to compressing large point-clouds during
offload from on-device sensors to perception services on a nearby
edge. We present the preliminary results from an algorithm to
demonstrate the opportunities from new
interpolation techniques. Our ongoing work focuses on further
improvements and comprehensive evaluation of the impact on end-to-end
perception of this approach. 


\section{Acknowledgment}
This work is supported by Ericsson Research Santa Clara.